\documentclass[11pt,letterpaper]{article}
\usepackage{ijcnlp2017}
\usepackage{graphicx}
\usepackage{times}
\usepackage{latexsym} 

\ijcnlpfinalcopy

\def\ijcnlppaperid{186}

\title{Recognizing Explicit and Implicit Hate Speech Using a Weakly Supervised Two-path Bootstrapping Approach}
\author{Lei Gao \\
  Texas A\&M University   \\ 
  {\tt sjtuprog@tamu.edu} \\\And
  Alexis Kuppersmith \\ 
  Stanford University \\
  {\tt Lex54@stanford.edu } \\\And
  Ruihong Huang \\
  Texas A\&M University   \\ 
  {\tt huangrh@cse.tamu.edu} \\}
\begin{document}

\maketitle

\begin{abstract}
  In the wake of a polarizing election, social media is laden with hateful content. To address various limitations of supervised hate speech classification methods including corpus bias and huge cost of annotation, we propose a weakly supervised two-path bootstrapping approach for an online hate speech detection model leveraging large-scale unlabeled data. This system significantly outperforms hate speech detection systems that are trained in a supervised manner using manually annotated data. Applying this model on a large quantity of tweets collected before, after, and on election day reveals motivations and patterns of inflammatory language.  
\end{abstract}

\section{Introduction}
Following a turbulent election season, 2016's digital footprint is awash with hate speech. Apart from censorship, the goals of enabling computers to understand inflammatory language are many. Sensing increased proliferation of hate speech can elucidate public opinion surrounding polarizing events. Identifying hateful declarations can bolster security in revealing individuals harboring malicious intentions towards specific groups. 

Recent studies on supervised methods for online hate speech detection \cite{waseem2016hateful,nobata2016abusive} have relied on 
manually annotated datasets, which are not only costly to create but also likely to be insufficient to obtain wide-coverage hate speech detection systems. 
This is mainly because online hate speech is relatively infrequent (among large amounts of online contents) and tends to transform rapidly following a new “trigger” event. 
Our pilot annotation experiment with 5,000 randomly selected tweets shows that around 0.6\% (31 tweets) of tweets are hateful.
The mass-scale (Yahoo! Finance online comments) hate speech annotation effort from Yahoo! \cite{nobata2016abusive} revealed that only 5.9\% of online comments contained hate speech.
Therefore, large amounts of online texts need to be annotated to adequately identify hate speech.
In recent studies \cite{waseem2016hateful,kwok2013locate}, the data selection methods and annotations are often biased towards a specific type of hate speech or hate speech generated in certain scenarios in order to increase the ratio of hate speech content in the annotated data sets, which 
however made the resulting annotations too distorted to reflect the true distribution of hate speech.
Furthermore, inflammatory language changes dramatically following new hate ``trigger'' events, which will significantly devalue annotated data. 

To address the various limitations of supervised hate speech detection methods, we present a weakly supervised two-path bootstrapping approach for online hate speech detection that requires minimal human supervision and can be easily retrained and adapted to capture new types of inflammatory language. Our two-path bootstrapping architecture consists of two learning components, an explicit slur term learner and a neural net classifier (LSTMs \cite{hochreiter1997long}), that can capture both explicit and implicit phrasings of online hate speech. 

Specifically, our bootstrapping system starts with automatically labeled online hateful content that are identified by matching a large collection of unlabeled online content with several hateful slur terms. Then two learning components will be initiated simultaneously.
 A slur term learner will learn additional hateful slur terms from the automatically identified hateful content. Meanwhile, a neural net classifier will be trained using the automatically labeled hateful content as positive instances and randomly sampled online content as negative instances. Next, both string matching with the newly learned slur terms and the trained neural net classifier will be used to 
recognize new hateful content from the large unlabeled collection of online contents. 
Then the newly identified hateful content by each of the two learning components will be used to augment the initially identified hateful content, which will be used to learn more slur terms and retrain the classifier. The whole process iterates. 
The design of the two-path bootstrapping system is mainly motivated to capture both explicit and implicit inflammatory language. 
Explicit hate speech is easily identifiable by recognizing a 
clearly hateful word or phrase. For example: 
    
  \vspace{.05in}
\noindent (1) {\it Don't talk to me from an anonymous account you faggot coward, whither up and die.}

\vspace{.05in}
\noindent (2) {\it And that's the kind of people who support Trump! Subhumans!}

\noindent In contrast, implicit hate speech employs circumlocution, metaphor, or stereotypes to convey hatred of a particular group, in which hatefulness can be captured by understanding its overall compositional meanings, For example:

\vspace{.05in}
\noindent (3) {\it Hillary's welfare army doesn't really want jobs. They want more freebies.}

\vspace{.05in}
\noindent (4) {\it Affirmative action means we get affirmatively second rate doctors and other professionals.}

Furthermore, our learning architecture has a flavor of 
co-training \cite{blum1998combining} in maintaining two learning components that concentrate on different properties of inflammatory language. By modeling distinct aspects of online hate speech, such a learning system is better equipped to combat semantic drift, which often occurs in self-learning where the learned model drifts away from the esteemed track. Moreover, training two complementary models simultaneously and utilizing both models to identify hate speech of different properties in each iteration of the learning process is important to maintain the learning momentum and to generate models with wide coverage. Indeed, our experimental results have shown that the two-path bootstrapping system is able to jointly identify many more hate speech texts (214,997 v.s 52,958 v.s 112,535) with a significantly higher F-score (48.9\% v.s 19.7\% v.s 26.1\%), when compared to the bootstrapping systems with only the slur term learner and only the neural net classifier. In addition, the evaluation shows that the two-path bootstrapping system identifies 4.4 times more hateful texts than hate speech detection systems that are trained using manually annotated data in a supervised manner. 

\section{Related Work} 

Previous studies on hate speech recognition
mostly used supervised approaches. Due to the sparsity of hate speech overall in reality, the data selection methods and annotations are often biased towards a specific type of hate speech or hate speech generated in certain scenarios. For instance, 
\citet{razavi2010offensive} conducted their experiments on 1525 annotated sentences from a company's log file and a certain newsgroup. 
\citet{warner2012detecting} 
labeled around $9000$ human labeled paragraphs from Yahoo!'s news group post and American Jewish Congress's website, and the labeling is restricted to anti-Semitic hate speech.
\citet{sood2012profanity} studied use of profanity on a dataset of 6,500 labeled comments from Yahoo! Buzz.
\citet{kwok2013locate} built a balanced corpus of 24582 tweets consisting of anti-black and non-anti black tweets. 
The tweets were manually selected from Twitter accounts that were believed to be racist based upon their reactions to anti-Obama articles.
\citet{burnap2014hate} collected hateful tweets related to the murder of Drummer Lee Rigby in 2013.
\citet{waseem2016hateful} 
collected tweets 
using hateful slurs, specific hashtags as well as suspicious user IDs.
Consequently, all of the 1,972 racist tweets are by 9 users, and the majority of sexist tweets are related to an Australian TV show.

\citet{djuric2015hate} is the first to study hate speech 
using a large-scale annotated data set. 
They have annotated 951,736 online comments from Yahoo!Finance, with 56,280 comments labeled as hateful.
\citet{nobata2016abusive} followed \citet{djuric2015hate}'s work. In addition to the Yahoo!Finance annotated comments,
they also annotated 1,390,774 comments from Yahoo!News. 
Comments in both data sets were randomly sampled from their corresponding websites with a focus on comments by users who were reported to have posted hateful comments.
We instead aim to detect hate speech w.r.t. its real distribution, 
using a weakly supervised method that does not rely on large amounts of annotations.

The commonly used classification methods in previous studies are logistic regression and Naive Bayes  classifiers. 
\citet{djuric2015hate} and \citet{nobata2016abusive} applied neural network models 
for training word embeddings, which were further used as features in a 
logistic regression model for classification.
We will instead train a neural net classifier  \cite{kim2014convolutional,lai2015recurrent,zhou2015c} 
in a weakly supervised manner in order to capture implicit and compositional hate speech expressions.


\citet{xiang2012detecting} is related to our research because they also used a bootstrapping method to discover offensive language from a large-scale Twitter corpus. However, their bootstrapping model is driven by mining hateful Twitter users, instead of content analysis of tweets as in our approach.  Furthermore, they recognize hateful Twitter users by detecting explicit hateful indicators (i.e., keywords) in their tweets while our bootstrapping system aim to detect both explicit and implicit expressions of online hate speech.

\section{The Two-path Bootstrapping System for Online Hate Speech Detection}

\subsection{Overview}
\begin{figure}[ht]
\centering
\includegraphics[width=7.6cm,height=10cm,keepaspectratio]{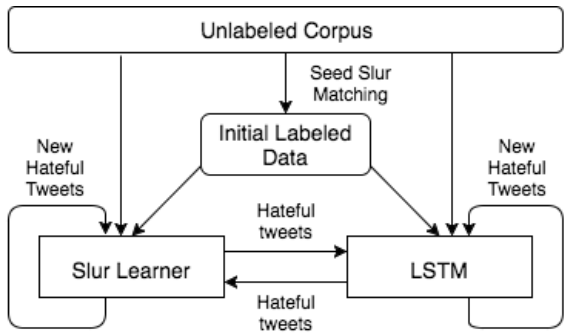}
\caption{Diagram of co-training model}\label{model}
  
\end{figure}

Figure \ref{model} illustrates that our 
weakly supervised hate speech detection system starts with
a few pre-identified slur terms as seeds and a large collection of unlabeled data instances. Specifically, we experiment with identifying hate speech from tweets. 
Hateful tweets will be 
automatically identified by matching the large collection of unlabeled tweets with slur term seeds. 
Tweets that contain one of the seed slur terms are labeled as hateful.

The two-path bootstrapping system consists of two learning components, an explicit slur term learner and a neural net classifier (LSTMs \cite{hochreiter1997long}), that can capture both explicit and implicit descriptions of online hate speech.
Using the initial seed slur term labeled hateful tweets, 
the two learning components will be initiated simultaneously.
The slur term learner will continue to learn additional hateful slur terms.
Meanwhile, the neural net classifier will be trained using the automatically labeled hateful tweets as positive instances and randomly sampled tweets as negative instances.
Next, both 
the newly learned slur terms and the trained neural net classifier will be used to 
identify new hateful content from the unlabeled large collection of tweets. 
 The newly labeled hateful tweets by each of the two learning components will be used to augment the initial slur term seed identified hateful tweet collection, which will be used to learn more slur terms and retrain the classifier in the next iteration.
The whole process then iterates. 

After each iteration, we have to determine if a stopping criterion is met and we should terminate the bootstrapping process. In general, a tuned threshold score is applied or a small annotated dataset is used to evaluate the learned classifiers. 
We adopt the latter method.
Specifically, the bootstrapping system stops when the precision of the LSTM classifier is lower than $0.6$ when evaluated using an existing small annotated tweet set \cite{waseem2016hateful}. 

\subsection{Automatic Data Labeling of Initial Data}

Seeing a hate slur term in a tweet strongly indicates that the tweet is hateful. Therefore, 
we use 20 manually selected slur terms to match with a large unlabeled tweet collection in order to quickly construct the initial small set of hateful tweets.  
Table \ref{seeds} shows the $20$ seed slurs we used.
\begin{table}[ht]
\begin{center}
\scalebox{0.9}{
\begin{tabular}{ l l l l l}
\hline
bimbo & chink & commie & coon & cunt \\
fag & faggot & feminazi & honky & islamist \\
libtard & muzzie & negro & nigger & paki \\
skank & subhuman & tranny & twat & wanker \\
\hline
\end{tabular}}
\end{center}
\caption{Seed slurs }\label{seeds}
\end{table}

We obtained our initial list of slurs from Hatebase\footnote{https://www.hatebase.org}, the Racial Slurs Database \footnote{http://www.rsdb.org}, and a page of LGBT slang terms\footnote{https://en.wikipedia.org/wiki/List\_of\_LGBT\_slang\_terms}. We ranked the slur terms by their frequencies in tweets, eliminating ambiguous and outdated terms. 
The slur "gypsy", for example, refers to derogatorily to 
people of Roma descent, but currently in popular usage is an idealization of a trendy bohemian lifestyle. 
The word "bitch" is ambiguous, sometimes a sexist slur but other times innocuously self-referential or even friendly. 

For these reasons, we only selected the top $20$ terms we considered reliable (shown in Table \ref{seeds}). We use both the singular and the plural form for each of these seed slur terms.


\subsection{Slur Term Learner}
The slur term learning component extracts individual words 
from a set of hateful tweets as new slurs.
Intuitively, if a word 
occurs significantly more frequently in hateful tweets than in randomly selected tweets, this term is more likely to be a hateful slur term. 
Following this intuition, we assign a score to each unique unigram 
that appears $10$ or more times in hateful tweets, and the score is calculated as the relative ratio of its frequency in the labeled hateful tweets over its frequency in the unlabeled set of tweets.
Then the slur term learner recognizes a unigram 
with a score higher than a certain threshold as a new slur. 
Specifically, we use the threshold score of $100$ in identifying individual word slur terms.

The newly identified slur terms will be used to match with 
unlabeled tweets in order to identify additional hateful tweets. A tweet that contains one of the slur terms is deemed to be a hateful tweet.

While we were aware of other more sophisticated machine learning models, one purpose of this research is to detect and learn new slur terms from constantly generated user data. Therefore, 
the simple and clean string matching based slur learner is designed to attentively look for specific words 
that alone can indicate hate speech. In addition, this is in contrast with the second learning component that uses a whole tweet and model its 
compositional meanings in order to recognize implicit hate speech. These two learners are complementary in the two-path bootstrapping system. 


\subsection{The LSTM Classifier}

We aim to recognize implicit hate speech expressions and capture composite meanings of tweets using a sequence neural net classifier. Specifically, our LSTM classifier has a single layer of LSTM units. The output dimension size of the LSTM layer is $100$. A sigmoid layer is built on the top of the LSTM layer to generate predictions. The input dropout rate and recurrent state dropout rate are both set to $0.2$. In each iteration of the bootstrapping process, the training of the LSTM classifier runs for $10$ epochs.

The input to our LSTM classifier is a sequence of words. 
We pre-process and normalize tokens in tweets following the steps suggested in \cite{pennington2014glove}.
In addition, we used the pre-processing of emoji and smiley described in a preprocess tool \footnote{https://pypi.python.org/pypi/tweet-preprocessor/0.4.0}. 
Then we retrieve word vector representations from the downloaded\footnote{https://code.google.com/archive/p/word2vec/}  pre-trained word2vec embeddings \cite{mikolov2013distributed}.

The LSTM classifier is trained using the automatically labeled hateful tweets as positive instances and randomly sampled tweets as negative instances, with the ratio of POS:NEG as 1:10.
Then the classifier is used to identify additional hateful tweets from the large set of unlabeled tweets. The LSTM classifier will deem a tweet as hateful if the tweet receives a confidence score of $0.9$ or higher. 
Both the low POS:NEG ratio and the high confidence score are applied to increase the precision of the classifier in labeling hateful tweets and control semantic drift in the bootstrapping learning process. 
To further combat semantic drift, we applied weighted binary cross-entropy as the loss function in LSTM. 

\subsection{One vs. Two Learning Paths}
 
As shown in Figure \ref{model}, if we remove one of the two learning components, the two-path learning system will be 
reduced to a usual self-learning system with one single learning path. 
For instance, if we remove the LSTM classifier, the slur learner will learn new slur terms from initially seed labeled hateful tweets and then identify new hateful tweets by matching newly learned slurs with unlabeled tweets. The newly identified hateful tweets will be used to augment the initial hateful tweet collection and additional slur terms can be learned from the enlarged hateful tweet set. The process will iterates. 
However as shown later in the evaluation section, single-path variants of the proposed two-path learning system are unable to receive additional fresh hateful tweets identified by the other learning component and lose learning momentum quickly.  

\subsection{Tackling Semantic Drifts}
Semantic drift is the most challenging problem in distant supervision and bootstrapping. First of all, we argue that the proposed two-path bootstrapping system with two significantly different learning components is designed to reduce semantic drift. According to the co-training theory \cite{blum1998combining}, the more different the two components are, the better. In evaluation, we will show that such a system outperforms single-path bootstrapping systems. Furthermore, we have applied several strategies in controlling noise and imbalance of automatically labeled data,  e.g., 
the high frequency and the high relative frequency thresholds enforced in selecting hate slur terms, as well as  the low POS:NEG training sample ratio and the high confidence score of 0.9 used in selecting new data instances for the LSTM classifier. 


\section{Evaluations}






\subsection{Tweets Collection}



We randomly sampled 10 million tweets from 67 million tweets collected from Oct. 1st to Oct. 24th using Twitter API. These 10 million tweets were used as the unlabeled tweet set in bootstrapping learning.
Then we continued to collect 62 million tweets spanning from Oct.25th to Nov.15th, essentially two weeks before the US election day and one week after the election. The 62 million tweets will be used to evaluate the performance of the bootstrapped slur term learner and LSTM classifier.
The timestamps of all these tweets are converted into EST.
By using Twitter API, the collected tweets were randomly sampled to prevent a bias in the data set.


\subsection{Supervised Baselines}
We trained two supervised models using the 16 thousand annotated tweets that have been used in a recent study   \cite{waseem2016hateful}.
The annotations distinguish two types of hateful tweets, 
sexism and racism, but we 
merge both categories and only distinguish hateful from non-hateful tweets.

First, we train a traditional feature-based classification model using logistic regression (LR). We apply the same set of features as mentioned in \cite{waseem2016hateful}. The features include character-level bigrams, trigrams, and four-grams.

In addition, for direct comparisons, we train a LSTM model using the 16 thousand annotated tweets, using exactly the same settings as we use for the LSTM classifier in our two-path bootstrapping system.

\subsection{Evaluation Methods}
We apply both supervised classifiers and our weakly supervised hate speech detection systems to the 62 million tweets in order to identify hateful tweets that were posted before and after the US election day.  
We evaluate both precision and recall for both types of systems. Ideally, we can easily measure precision as well as recall for each system if we have ground truth labels for each tweet. 
However, it is impossible to obtain annotations for such a large set of tweets. The actual distribution of hateful tweets in the 62 million tweets is unknown. 

Instead, to evaluate each system, 
we randomly sampled 1,000 tweets from the whole set of hateful tweets that {\it had been  tagged as hateful} by the corresponding system. Then we annotate the sampled tweets and use them to estimate precision and recall of the system. In this case, 
\[ precision = \frac{n}{ 1000 } \]
\[ recall \propto precision \cdot N \]

Here, $n$ refers to the number of hateful tweets that human annotators identified in the 1,000 sampled tweets, and $N$ refers to the total number of hateful tweets the system tagged in the 62 million tweets. 
We further calculated system recall by normalizing the product, $precision \cdot N$, with an estimated 
total number of hateful tweets that exist in the 62 million tweets, which was obtained by  
multiplying the estimated hateful tweet rate of 0.6\%\footnote{We annotated 5,000 tweets that were randomly sampled during election time and 31 of them were labeled as hateful, therefore the estimated hateful tweet rate is 0.6\% (31/5,000).} with the exact number of tweets in the test set. 
Finally, we calculate F-score using the calculated recall and precision.

Consistent across the statistical classifiers including both logistic regression classifiers and LSTM models, only tweets that receive a confidence score over $0.9$ were tagged as hateful tweets.  

\begin{table*}[ht]
\begin{center}
\scalebox{0.94}{
\begin{tabular}{|l|c|c|c|c|c|}
\hline \bf Classifier & \bf Precision & \bf Recall & \bf F1 & \bf \# of Predicted Tweets & \bf \# of Estimated Hateful  \\ \hline
\multicolumn{6}{|c|}{Supervised Baselines} \\ \hline
Logistic Regression & 0.088 & 0.328& 0.139& \bf{1,380,825} & 121,512\\ 
LSTMs & \bf{0.791} & 0.132& 0.228&  62,226 & 49,221 \\ \hline
\multicolumn{6}{|c|}{The Two-path Weakly Supervised Learning System} \\ \hline
LSTMs         & 0.419 & 0.546& 0.474& 483,298 &  202,521  \\ 
Slur Matching  & 0.565 & 0.398& 0.468 &  261,183 &  147,595 \\ 
Union & 0.422 & \bf{0.580} & \bf{0.489} & 509,897 & \bf{214,997}\\ \hline
Union* & 0.626* & 0.258* & 0.365* & - & - \\ \hline
\multicolumn{6}{|c|}{Variations  of the Two-path Weakly Supervised Learning System} \\ \hline
Slur Matching Only  &  0.318 & 0.143 & 0.197& 166,535 & 52,958 \\ 
LSTMs Only  &  0.229  & 0.303 & 0.261 & 491,421 & 112,535 \\ 
\hline 
\end{tabular}
}
\end{center}
\caption{\label{pm} Performance of Different Models }
\end{table*}

\subsection{Human Annotations}
When we annotate system predicted tweet samples, we essentially adopt the same definition of hate speech as used in \cite{waseem2016hateful}, which considers tweets that explicitly or implicitly propagate stereotypes targeting a specific group whether it is the initial expression or a meta-expression discussing the hate speech itself (i.e. a paraphrase). 
In order to ensure our annotators have a complete understanding of online hate speech, we asked two annotators to first discuss over a very detailed annotation guideline of hate speech, then
 annotate separately. This went for several iterations. 

Then we asked the two annotators to annotate the 1,000 tweets that were randomly sampled from all the tweets tagged as hateful by the supervised LSTM classifier. 
The two annotators reached an inter-agreement  Kappa \cite{cohen1960coefficient} score of 85.5\%. 
Because one of the annotators become unavailable later in the project,  the other annotator annotated the remaining sampled tweets.

\subsection{Experimental Results} 

\noindent {\bf Supervised Baselines}

The first section of Table \ref{pm} shows the performance of the two supervised models when applied to 62 million tweets collected around election time. 
We can see that the logistic regression model suffers from an extremely low precision, which is less than 10\%. While this classifier aggressively labeled a large number of tweets as hateful, only 121,512 tweets are estimated to be truly hateful.
In contrast, the supervised LSTM classifier 
has a high precision of around 79\%, however, this classifier is too conservative and only labeled a small set of tweets as hateful. 

\begin{table}[t]
\begin{center}
\begin{tabular}{|l|r|r|r|r|}
\hline \bf Its & \bf Prev & \bf Slur Match  &  \bf LSTMs     \\ \hline
1   & 8,866 & 422 &  3,490   \\ 
2  & 12,776 &  4,890 &  13,970 \\ 
3   &  27,274 & 6,299  & 21,579 \\
4   & 50,721 & 9,895 &  22,768 \\  
\hline
\end{tabular}
\end{center}

\caption{Number of Labeled Tweets in Each Iteration }\label{boot}
\end{table}

\noindent {\bf The Two-path Bootstrapping System}

Next, we evaluate our weakly supervised classifiers which were obtained using only $20$ 
seed slur terms and a large set of unlabeled tweets.
The two-path weakly supervised bootstrapping system ran for four iterations. 
The second section of Table \ref{pm} shows the results for the two-path weakly supervised system. 
The first two rows show the evaluation results for each of the two learning components in the two-path system, the LSTM classifier and the slur learner, respectively. 
The third row shows the results for the full system.
We can see that the full system {\bf Union}  is significantly better than the supervised LSTM model in terms of recall and F-score. 
Furthermore, we can see that a significant portion of hateful tweets were identified by both components and the weakly supervised LSTM classifier is especially capable to identify a large number of hateful tweets.  
Then the slur matching component obtains an precision of around 56.5\% and can identify roughly 3 times of hateful tweets compared with the supervised LSTM classifier.
The last column of this section shows the performance of our model on a collection of human annotated  tweets as introduced in the previous work \cite{waseem2016hateful}.
The recall is rather low because the data we used to train our model is quite different from this dataset which contains tweets related to a TV show \cite{waseem2016hateful}. 
The precision is 
only slightly lower than previous supervised models that were trained using the same dataset. 

Table \ref{boot} shows the number of hateful tweets our bootstrapping system identified in each iteration during training.
Specifically, 
the columns {\bf Slur Match} and {\bf LSTMs} show 
the number of hateful tweets identified by the slur learning component and the weakly supervised LSTM classifier respectively. We can see that both learning components steadily label new hateful tweets in each iteration and the LSTM classifier often labels more tweets as hateful compared to slur matching. 

\begin{table}[t]
\begin{center}
\begin{tabular}{|c|c|c|}
\hline  \bf Intersection & \bf LSTM Only  & \bf Slur Only   \\ \hline
 234,584 & 248,714 & 26,599 \\  
\hline
\end{tabular}
\end{center}

\caption{Number of Hateful Tweets in Each Segment}
\label{perf}
\end{table}

Furthermore, we found that many tweets were labeled as hateful by both
slur matching and the LSTM classifier. Table \ref{perf} shows the number of hateful tweets in each of the three segments, hateful
tweets that have been labeled by both components as well
as hateful tweets that were labeled by one component only.
Note that the three segments of tweets are mutually exclusive
from others. We can see that many tweets were labeled
by both components and each component separately labeled
some additional tweets as well. This demonstrates that hateful tweets
often contain both explicit hate indicator phrases and implicit
expressions. Therefore in our two-path bootstrapping system, the
hateful tweets identified by slur matching are useful for improving
the LSTM classifier, vice versa. This also explains
why our two-path bootstrapping system learn well to
identify varieties of hate speech expressions in practice.

\noindent {\bf One-path Bootstrapping System Variants}

In order to understand how necessary it is to 
maintain two learning paths for online hate speech detection, we also ran two experiments with one learning component removed from the loop each time. 
Therefore, the reduced bootstrapping systems can only repeatedly learn 
explicit hate speech (with the slur learner) or implicit hateful expressions (with the LSTM classifier). 

The third section of Table \ref{pm} shows the evaluation results of the two single-path variants of the weakly supervised system. 
We can see that both the estimated precision, recall, F score and the estimated number of truly hateful tweets by the two systems are significantly lower than the complete two-path bootstrapping system, which suggests that our two-path learning system can effectively capture diverse descriptions of online hate speech, maintain learning momentums as well as effectively combat with noise in online texts.

\section{Analysis}

\subsection{Analysis of the Learned Hate Indicators}
\vspace{.1in}
\begin{table}[ht]
\begin{center}
\scalebox{0.9}{
\begin{tabular}{ l l l l}
\hline
berk & chavs & degenerates & douches\\
facist & hag & heretics & jihadists\\ 
lesbo & pendejo & paedo & pinche\\
retards & satanist & scum & scumbag\\
slutty & tards & unamerican & wench\\
\hline
\end{tabular}}
\end{center}
\caption{New slurs learned by our model}\label{new slur}
\end{table}
\vspace{-.1in}
We have learned 306 unigram phrases using the slur term learning component.  
Among them, only 45 phrases were seen in existing hate slur databases while the other terms, 261 phrases in total, were only identified in real-world tweets. 
Table \ref{new slur} shows some of the newly discovered hate indicating phrases.
Our analysis shows that 86 of the newly discovered hate indicators are strong hate slur terms 
and the remaining 175 indicators are related to discussions of identity and politics such as 'supremacist' and 'Zionism'.

\subsection{Analysis of LSTM Identified Hateful Tweets}

The LSTM labeled 483,298 tweets as hateful, and 172,137 of them do not contain any of the original seed slurs or our learned indicator phrases. The following are example hateful tweets 
that have no explicit hate indicator phrase:

\vspace{.05in}
\noindent (1) {\it @janh2h The issue is that internationalists keep telling outsiders that they're just as entitled to the privileges of the tribe as insiders.}




\vspace{.05in}
\noindent (2) {\it This is disgusting! Christians are very tolerant people but Muslims are looking to wipe us our and dominate us! Sen https://t.co/7DMTIrOLyw} 

We can see that the hatefulness of these tweets is determined by their overall compositional meanings 
rather than a hate-indicating slur.

\subsection{Error Analysis}
The error of our model comes from semantic drift in bootstrapping learning, which partially results from the complexity and dynamics of language. Specifically, we found dynamic word sense of slurs and natural drifting of word semantic.
Many slur terms are ambiguous and have multiple word senses. For instance, ``Chink'', an anti-Asian epithet, can also refer to a patch of light from a small aperture. Similarly, ``Negro'' is a toponym in addition to a racial slur. Further, certain communities have reclaimed slur words. Though the word ``dyke'' is derogatory towards lesbians, for example, some use it self-referentially to destigmatize it, a phenomenon we sometimes encountered. 

 
\subsection{Temporal Distributions of Tagged Hateful Tweets}
By applying our co-training model on the 62 million tweets
corpus, we found around 510 thousand tweets labeled as hateful
in total. 

\begin{figure}[ht]
  \centering
  \includegraphics[width=7.6cm,keepaspectratio]{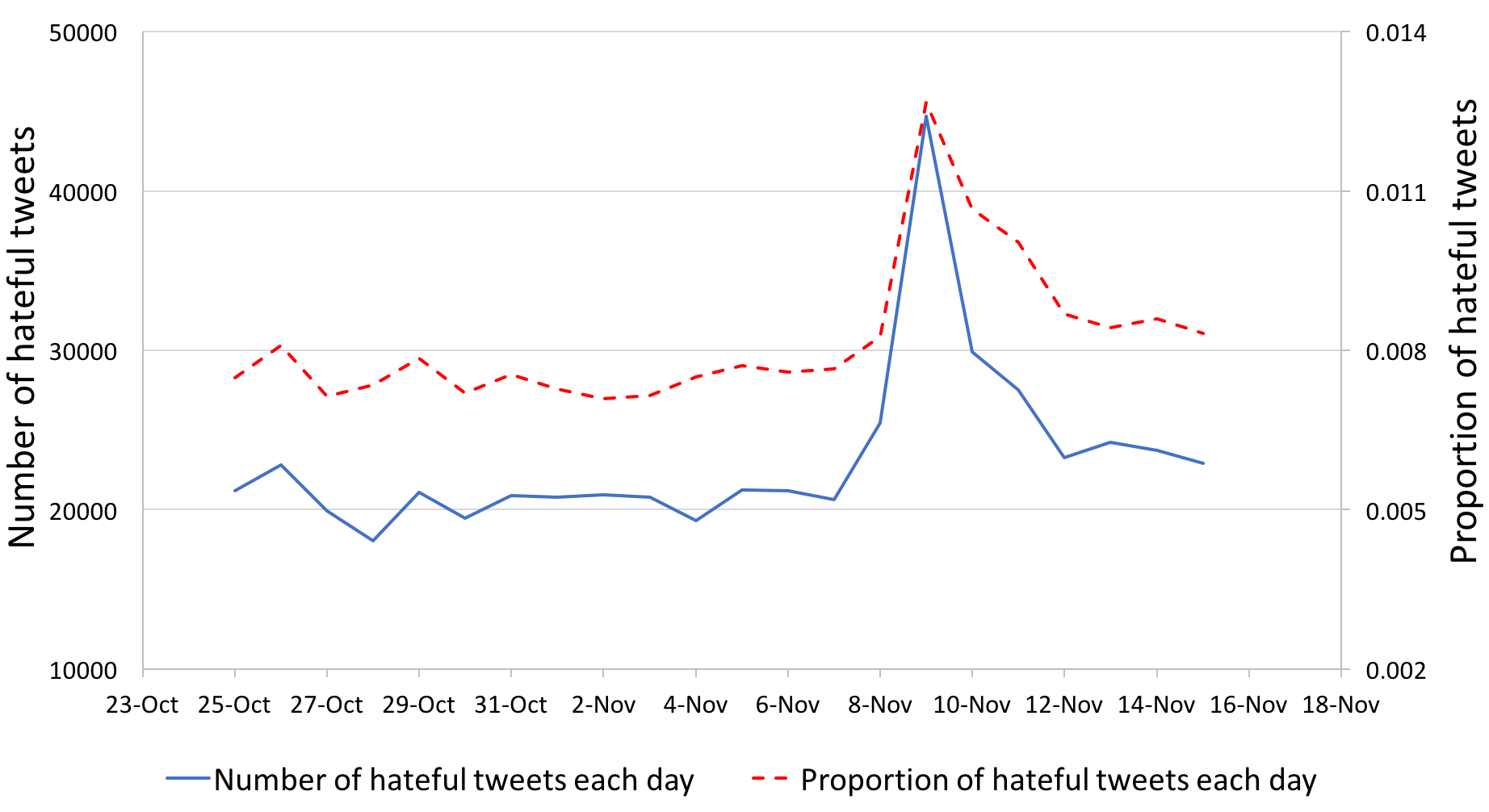}
\caption{Temporal Distribution of Hateful Tweets}
\label{trend}
\end{figure}
The figure \ref{trend} displays the temporal distribution of hateful tweets. There is a spike in hateful tweets from Nov.7th to Nov.12th in terms of both number of hateful tweets and ratio of hateful tweets to total tweets.

\subsection{Most Frequent Mentions and Hashtags of Tagged Hateful Tweets}

Table \ref{mention} and \ref{hashtag} show the top 30 most frequent mentions 
in hateful tweets. They are ranked by frequency from left to right and from top to bottom. 

It is clear that the majority of mentions found in tweets tagged as hateful address polarizing political figures (i.e. @realDonaldTrump and @HillaryClinton), indicating that hate speech is often fueled by partisan warfare. Other common mentions include news sources, such as Politico and MSNBC, which further support that "trigger" events in the news can generate inflammatory responses among Twitter users. Certain individual Twitter users also received a sizable number of mentions. @mitchellvii is a conservative activist whose tweets lend unyielding support to Donald Trump. Meanwhile, Twitter user @purplhaze42 is a self-proclaimed anti-racist and anti-Zionist. Both figured among the most popular recipients of inflammatory language. 

Table \ref{hashtag} shows that the majority of hashtags also indicate the political impetus behind hate speech with hashtags such as \#Trump and \#MAGA (Make America Great Again, Trump's campaign slogan) among the most frequent. 
The specific televised events also engender proportionally large amounts of hateful language as they can be commonly experienced by all television-owning Americans and therefore a widely available target for hateful messages. 

\begin{table}[ht]
\begin{center}
\scalebox{0.8}{
\begin{tabular}{ l l l}
\hline
@realDonaldTrump & @HillaryClinton & @megynkelly \\
@CNN & @FoxNews & @newtgingrich \\
@nytimes & @YouTube & @POTUS\\
@KellyannePolls & @MSNBC & @seanhannity \\
@washingtonpost & @narendramodi & @CNNPolitics \\
@PrisonPlanet & @guardian & @JoyAnnReid \\
@BarackObama & @thehill & @BreitbartNews \\
@politico & @ABC & @AnnCoulter \\
@jaketapper & @ArvindKejriwal & @FBI \\ 
@mitchellvii & @purplhaze42 & @SpeakerRyan \\
\hline
\end{tabular}}
\end{center}
\caption{List of Top 30 Mentions in Hateful Tweets During Election Days}\label{mention}
\end{table}

\begin{table}[ht]
\begin{center}
\scalebox{0.8}{
\begin{tabular}{ l l l}
\hline
\#Trump  &  \#ElectionNight &   \#Election2016    \\
\#MAGA  &   \#trndnl  &  \#photo   \\
 \#nowplaying  &  \#Vocab  &   \#NotMyPresident    \\
 \#ElectionDay &    \#trump    & \#ImWithHer    \\
 \#halloween &   \#cdnpoli    & \#Latin    \\
  \#Hillary  &  \#WorldSeries &  \#1    \\
 \#Brexit  &   \#Spanish    & \#auspol     \\
 \#notmypresident  &  \#C51 &  \#NeverTrump    \\
\#hiring &    \#bbcqt     & \#USElection2016    \\
\#tcot &    \#TrumpProtest   & \#XFactor   \\

\hline
\end{tabular}}
\end{center}
\caption{List of Top 30 Hashtags in Hateful Tweets During Election Days}\label{hashtag}
\end{table}

\vspace{-0.1in}
\section{Conclusions}
Our work focuses on the need to capture both explicit and implicit
hate speech from an unbiased corpus. To address these issues, we proposed a weakly supervised two-path bootstrapping model to identify hateful language in randomly sampled tweets. Starting from 20 seed rules, we found 210 thousand hateful tweets from 62 million tweets collected during the election. 
Our analysis shows a strong correlation between temporal 
distributions of hateful tweets and the election time,
as well as the partisan impetus behind large amounts of inflammatory language. 
In the future, we will look into linguistic phenomena that 
often occur in hate speech, such as sarcasm and humor, to further improve hate speech detection performance. 

\bibliography{ijcnlp2017}
\bibliographystyle{ijcnlp2017}

\end{document}